\pdfoutput=1 

\documentclass[AMA,STIX1COL]{WileyNJD-v2}

\articletype{Article Type}%

\received{}
\revised{}
\accepted{25 Oct 2021}

\usepackage{multirow}
\usepackage{booktabs}
\usepackage{paralist}
\usepackage{lipsum}

\newcommand{\xhdr}[1]{\vspace{3pt}\noindent\textbf{#1} }

\newcommand{\refattsmall}{J-Att}
\newcommand{\errorcam}{ErrMap}
\newcommand{\inspector}{Justifying}

\newcommand{\vqasystem}{VQA-Net}
\newcommand{\bertattn}{AT-BERT}
\newcommand{\usability}{\emph{Helpfulness}}
\newcommand{\relevance}{\emph{Relevance}}
\newcommand{\relsmall}{REL}
\newcommand{\helpsmall}{HELP}

\newcommand{\changed}[1]{\textcolor{black}{#1}}

\newcommand{\reva}[1]{\textcolor{black}{#1}}
\newcommand{\revb}[1]{\textcolor{black}{#1}}

\begin{document}

\title{Generating and Evaluating Explanations of Attended and Error-Inducing Input Regions for VQA Models}

\author[1]{Arijit Ray*}

\author[1]{Michael Cogswell}

\author[1]{Xiao Lin}

\author[2]{Kamran Alipour}

\author[1]{Ajay Divakaran}

\author[1]{Yi Yao}

\author[1]{Giedrius Burachas}

\authormark{Arijit Ray \textsc{et al}}

\address[1]{\orgdiv{Center for Vision Technologies}, \orgname{SRI International}, \orgaddress{\state{Princeton, NJ}, \country{USA}}}

\address[2]{\orgdiv{Department of Computer Science}, \orgname{University of California, San Diego}, \orgaddress{\state{La Jolla, CA}, \country{USA}}}

\corres{*Arijit Ray, \email{arijit.ray93@gmail.com}}

\abstract[Summary]{Attention maps, a popular heatmap-based explanation method for Visual Question Answering (VQA), are supposed to help users understand the model by highlighting portions of the image/question used by the model to infer answers. 
However, we see that users are often misled by current attention map visualizations that point to relevant regions despite the model producing an incorrect answer.
Hence, we propose Error Maps that clarify the error by highlighting image regions where the model is prone to err. Error maps can indicate when a correctly attended region may be processed incorrectly leading to an incorrect answer, and hence, improve users' understanding of those cases.
To evaluate our new explanations, we further introduce a metric that simulates users' interpretation of explanations to evaluate their potential helpfulness to understand model correctness. 
We finally conduct user studies to see that our new explanations help users understand model correctness better than baselines by an expected 30\% and that our proxy helpfulness metrics correlate strongly ($\rho>0.97$) with how well users can predict model correctness.}


\maketitle

\footnotetext{Project page: \url{https://bit.ly/3cfl9Tc}}

\section{Introduction}

\revb{Good mental models of co-workers are one of the important factors for successful collaboration amongst humans \cite{engel2014reading, biran2017explanation, miller2019explanation}. 
Similarly, for human-machine collaboration, having a good mental model can help users calibrate their expectations on machines' robustness to establish trust of machine decisions \cite{bansal2019beyond}.}
To improve users' mental models of machines, there has been a recent focus on generating ``explanations'' that shed light on machines' inner workings \cite{gunning2019darpa}. 
In the context of \reva{Visual Question Answering} (VQA), visualizing an attention map over an image is a popular form of explanation that highlights relevant regions used by the system for answering the question \cite{anderson2018bottom, huk2018multimodal, yang2016stacked, lu2016hierarchical}.  

\begin{figure}[t]
\centering
\includegraphics[width=0.85\textwidth]{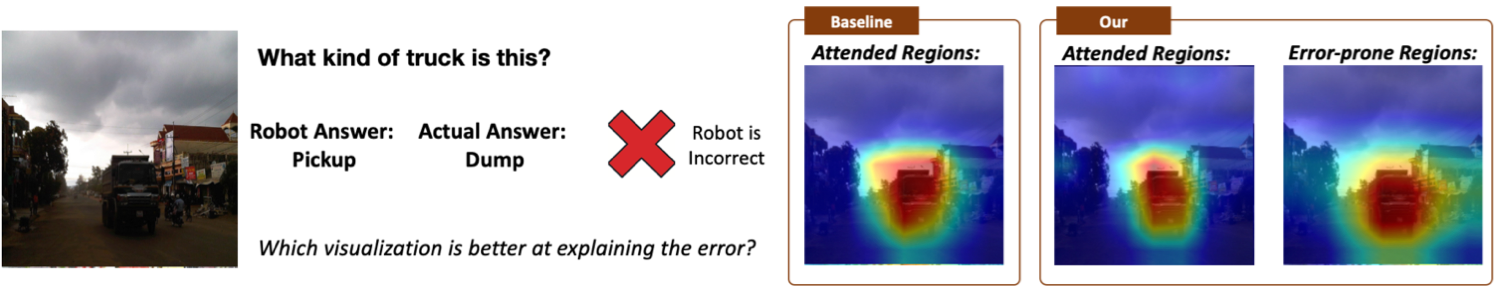}
\caption{We model how humans interpret heatmap explanations and propose novel explanations that are helpful to users for understanding system performance. We introduce error maps that point to regions that potentially cause system error. If error maps align with attended regions (as pointed to by attention maps), it indicates a possibility of system failure, and vice versa. 
} 
\label{fig:teaserfig}
\end{figure}
For example, as shown in Figure \ref{fig:teaserfig}, to explain the answer ``pickup'' produced by a VQA system to the question ``What kind of truck is this?'', an attention map is generated to highlight the vehicle region.

Users' ability to understand \reva{a} VQA model's performance while using such explanations is usually tested by asking users to predict model correctness while viewing the explanations along with the input question and image  \cite{chandrasekaran2017takes, alipour2020study}. However, in these studies, attention maps haven't been shown to be very helpful \cite{chandrasekaran2017takes, ray2019can, alipour2020study, hendricks2016generating} in assisting users to predict VQA system correctness. 

To analyze how humans interpret explanations and why current attention maps don't help, we first observe that users generally think a VQA model will produce a correct answer when the attention maps point to the relevant evidence in the image (measured using human attention as described in Section \ref{sec:helpfulness} and illustrated in Figure \ref{fig:atten_rel_correct}). However, as shown in Figure \ref{fig:teaserfig}, we frequently observe that attention maps point to relevant regions (the truck) even when VQA answers incorrectly (``pickup'' instead of ``dump'').
This is misleading to a user who would intuitively guess that the system will be correct because it looked at the right locations.

To clarify cases where an incorrect answer is produced in spite of attention pointing to the relevant regions, we consider an additional mode of explanation to highlight image areas that may be processed incorrectly - we refer to this as an Error Map. Taking Figure \ref{fig:teaserfig} for example, the Error Map points to the relevant vehicle region indicating that the vehicle may not be processed correctly (maybe because of low illumination, complexity of the entity, similarity to confusing entities, etc.). 
We generate error maps by training a \inspector\ module to learn model correctness and then using GradCAM \cite{selvaraju2017grad} to highlight visual evidence that contribute to a prediction of model error.

In order to measure the potential of our explanations to help users predict model correctness, we introduce a proxy helpfulness metric by simulating the way humans interpret them. The metric is based on 1) judging whether the explanation highlights information the human thinks is relevant and 2) using the relevance as an indicator to decide if the system can produce the correct answer or not. 
Attention maps that point to relevant regions for correct answers would be helpful according to such a metric. We also outline a few ways to optimize the visualization of attention maps to maximize these helpfulness characteristics.

Finally, we demonstrate that our new error maps, combined with attention maps, can improve users' ability to predict model performance compared to using baseline attention maps or using no explanations.

In summary, our major contributions include:
1) We introduce an  \textbf{automated proxy metric to evaluate the transparency of attention and error maps}. Although user studies are still the gold standard to evaluate attention/error maps, \textbf{our proxy metric can help reduce the need for multiple expensive and laborious user studies during development.}
2) We propose a novel mode of explanation, \textbf{Error Maps, that infers the regions that might induce error in the output}. We \textbf{demonstrate the effectiveness} of error maps, combined with attention maps, in improving helpfulness of explanations from the perspective of assisting users to predict model correctness.
3) We employ our metrics to optimize the visualization of attention maps from multiple heads of transformers \cite{vaswani2017attention}, which otherwise is done via manual inspection in literature.
4) We conduct \textbf{user studies to verify the effectiveness of our new attention and error maps} and show a \textbf{~30\% improvement} in user accuracy for predicting model performance when using helpful attention and error maps.

In the sections below, we first describe our VQA model and how attention maps are commonly generated in literature. We then describe how users interpret these attention maps, the problem with current attention maps, how error maps can help improve helpfulness. By simulating user interpretation, we describe our proxy metric for evaluating helpfulness of these attention/error maps. We then describe how we generate error maps and their evaluation.

\section{\vqasystem\ }
\label{sec:vqanet}
\begin{figure}[h]
\centering
\includegraphics[height=55px]{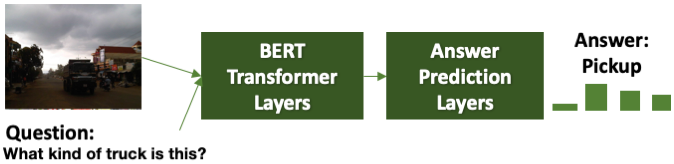}
\caption{High Level diagram of our VQA model} 
\label{fig:vqamodel}
\end{figure}
\changed{We first describe our AI system of choice - a Visual Question Answering (VQA) model, which we use to study what makes explanations helpful to make better explanations. }
We build a transformer-based \cite{vaswani2017attention} VQA system that answers questions about images. 

As shown in Figure \ref{fig:vqamodel}, our \vqasystem\ takes an image, $I$ and a question, $Q$ as input and outputs a probability distribution over $3129$ answer choices. 
The image $I$ is encoded into a $7\times7\times2048$ feature map using a Resnet152 \cite{he2016deep} and further encoded into a $49\times512$ input feature matrix, $H_m$.
We also extract $36$ object-based features from the image using a Region Proposal Network \cite{he2017mask}, which is encoded into a $36 \times 512$ object feature array, $H_o$. 
Finally, the question is encoded into a $30 \times 512$ feature array, $H_q$ with a max word length of 30 \footnote{Questions below that length are padded with 0's}.

We employ transformer-based attention layers (referred to as \textbf{\bertattn}) that jointly take in the image and question features. The input to the \bertattn\ layers are concatenated image and question features ($H_m$, $H_o$ and $H_q$). 
Hence, the input dimension to the \bertattn\ layers consists of $115$ ($30+36+49$) input tokens of dimension $512$ each.   
Our \bertattn\ has 4 layers with 12 heads each. In each head and layer of \bertattn, 115 input tokens generate an attention over all other 115 input tokens. Hence, a \bertattn\ module with $l$ layers and $h$ heads produces weights, $W$, of dimension $l\times h\times d \times d$ attention weights. $d$ is the number of input tokens, which is 115 in our case.
Finally, the \vqasystem\ predicts a softmax probability distribution over 3129 answer choices (i.e., the top VQA2.0 \cite{antol2015vqa} answers) from the attention weighted feature values. 
Our \vqasystem{} is trained on the VQA 2.0 dataset \cite{goyal2017making}.

\xhdr{Attention Map Generation} 
Following common practice in literature \cite{alipour2020impact}, we display attention by choosing the layers and heads according to a visual inspection on a held-out val-train set. Specifically, the attention is selected by averaging the attention weights on image features by all input tokens in all heads of the last layer of \bertattn. As we will see in the section below, this sort of attention map visualization does not help users to understand model correctness.

\section{Human Interpretation of Explanations}
\vspace{-2mm}
\begin{figure}[h]
\centering
\includegraphics[height=120px]{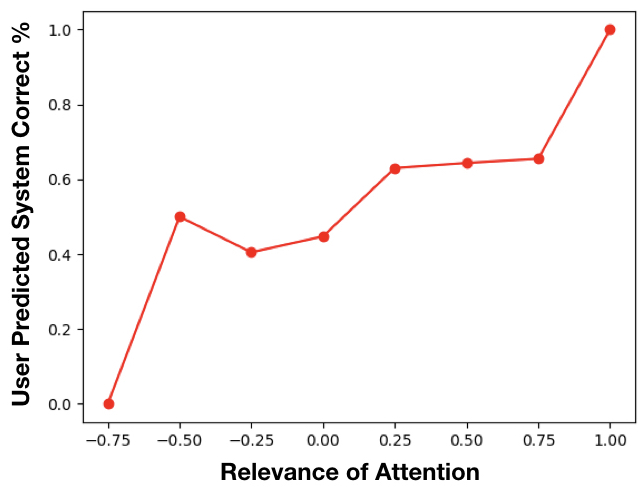}
\caption{Users believe a model will be correct when the attention points to relevant areas of the image.} 
\label{fig:atten_rel_correct}
\end{figure}
\vspace{-2mm}
\label{sec:helpfulness}
\changed{Our goal is to improve human's understanding of the model when seeing explanations  such as attention maps as described in the previous section. In order to test users' understanding of the model, we conduct a user-study to see how well users can predict the performance of the network when seeing these attention maps. 
Given an Image-Question (IQ) pair and an attention map, users needed to judge if a VQA model can produce the correct answer or not.  We compare this to a group of users seeing no explanations (only image and question) while trying to predict model correctness.
We provided a bonus incentive that was scaled to their accuracy in predicting the VQA performance.} 

\xhdr{Users get mislead by current attention maps}
We observe that the users' accuracy in predicting correctness does not improve over not seeing any attention maps. In fact, we see a slight decrease (57.18\% without explanations, 56.87\% when seeing attention maps). \reva{This suggests that the guessing accuracy in either case (with or without attention maps) is close to random guessing (50\%)}. To understand why we see no improvement in correctness prediction accuracy, we first analyze how humans interpret these explanations and the fault with the current visualization of the attention maps.

\xhdr{Users associate relevance with correctness}

We sample subsets of IQ pairs from our user study set and plot the average relevance of attention in those sets to the percentage of cases users think the model will be correct at to compute Figure \ref{fig:atten_rel_correct}.
We see that users predict the model will be correct more on average when attention maps point to ``relevant'' areas of the image.
Our definition of \textbf{``Relevance''} makes use of annotated human attentions 
for IQ pairs from the VQA \cite{antol2015vqa} dataset collected by \cite{vqahat}. This dataset of attention maps show where humans looked to answer VQA questions.
Using these human attention maps, we define the \relevance\, $\relsmall_A$ of an explanation $A$ as the Spearman rank correlation $\phi(\cdot)$ of the explanation with the human attention $A_h$ for a given IQ pair \footnotemark:
\vspace{-1mm}
\begin{align}
    \relsmall^A = \phi(A, A_h)
\end{align} \vspace{-2mm}
\footnotetext{$7\times7$ attention values are flattened into a vector.}

We choose a rank correlation over the commonly used Pearson's correlation because we want to measure whether the directions of change of attention values are similar to that of the human attention values rather than the absolute magnitudes. 
Absolute magnitudes depend on the temperature of the attention distribution, which is irrelevant to the similarity we intend to measure, yet can still change the Pearson correlation significantly. 

\begin{figure}[t]
\centering
\includegraphics[height=100px]{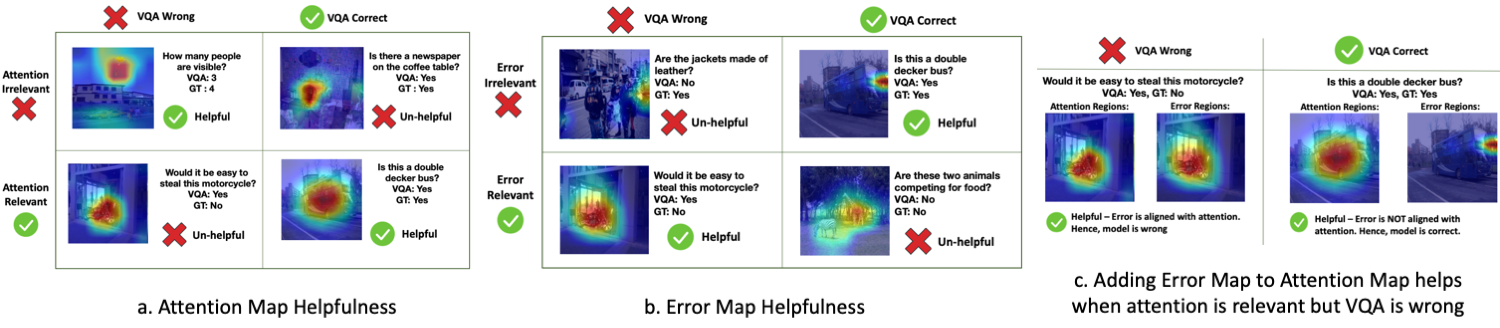}
\caption{\textbf{a.} Attention is perceived as helpful when it points to the relevant regions while the VQA model produces the correct answer (bottom-right quadrant) and points to the irrelevant regions while the VQA model produces incorrect answers (top-left quadrant).
\textbf{b.} Error map is perceived as helpful when it points to the irrelevant regions when the VQA model answers correctly (top-right quadrant) and points to the relevant regions when the VQA model answers incorrectly (bottom-left quadrant).
\textbf{c.} Combining attention and error maps improves the helpfulness of explanation especially for cases where attention points to the relevant regions while the VQA answers incorrectly (bottom-left quadrant of Figure \ref{fig:atten_help_grid}). With error map, the incorrect answer is better explained (left panel, this figure).
} 
\label{fig:atten_help_grid}
\end{figure}

 Using the fact that users associate relevance with correctness as seen in Figure \ref{fig:atten_rel_correct}, we can assume that users find it intuitive to understand model behavior when attention maps point to relevant regions when the system is correct and vice versa. 
Figure \ref{fig:atten_help_grid}a illustrates this point- it is intuitive to understand machine's correctness for the upper left quadrant (the attention is relevant and the machine is correct) and for the bottom right quadrant (the attention is irrelevant and the machine is incorrect). However if the attention is relevant and the answer is incorrect or vice versa (quadrants 2 and 3), it is unclear why the machine is correct/incorrect.

\xhdr{Attention Maps seem to be relevant even for incorrect cases}
Unfortunately, with current attention map visualizations, we observe that the attention maps often point to relevant image regions even when the model produces an incorrect answer (the counterintuitive case seen in Figure \ref{fig:atten_help_grid}a, quadrant 3). If we compute the average $REL^A$ for current attention maps for correct answer cases and incorrect answer cases, we see they are 0.32 and 0.31 respectively. This indicates that there is very little visual difference between the correct and incorrect cases. Hence, we need another mode of explanation that explains why the model was incorrect despite attention pointing to similarly relevant areas as correct cases.

\xhdr{How Error Maps can help}
\changed{To clarify cases where the model produces an incorrect answer even when the attention is relevant, we introduce a novel method of explanation - Error Maps, that indicate whether the attended region might be processed correctly or not.}
\changed{If users indeed use the relevance of the attention as a signal for when the model is correct, we can assume that users would similarly use the relevance of the error map to judge if the model will answer incorrectly or not.
If an error map points to relevant areas of the image or to the attended regions, a user would assume that those areas may not be processed correctly. 
As illustrated in Figure \ref{fig:atten_help_grid}b, it is easy to understand why a machine failed when the error map points to the relevant region needed to answer the question (bottom left quadrant) and why a machine succeeded if it points away from the relevant regions (upper right quadrant). }

\section{Evaluating Attention/Error Maps- A Proxy Helpfulness Measure}
\label{sec:helpfulness_approach}
\changed{Using our knowledge that humans judge model correctness based on how well the explanations point to relevant areas, we design a proxy helpfulness metric based on relevance to evaluate the potential of an attention map to help users to understand machine correctness. We can use this metric to evaluate explanations quickly during development without having to conduct repeated user-studies.
Note that, in this work, we focus on helpfulness of explanations rather than faithfulness of explanations to the model's inner workings.}



\xhdr{Helpfulness of Attention Maps:}
Our helpfulness metric for Attention Maps is higher if their relevance is higher for cases where the VQA model produces correct answers than it is for cases where the VQA model produces incorrect answers. The higher the difference, the better a user can visually tell apart when a model would be incorrect based on the attention relevance.  
As such, it is defined on a set $S$ of IQ pairs where sometimes the model is wrong and sometimes the model is right.

We conduct a z-test\footnotemark \footnotetext{z-test calculated using statsmodels.org Python package} for comparing the set of correct answer \emph{Relevances} to the same for incorrect answers. Higher the t-statistic of the z-test, the higher the significance of the difference and hence, the more the helpfulness. Let $\relsmall^A_c$ be the set of correct answers relevances and $\relsmall^A_w$ for wrong answers, we compute
\begin{align}
    \helpsmall^A_Z &= ztest(\relsmall^A_c, \relsmall^A_w)
\end{align}

\xhdr{This helpfulness metric indeed correlates to user accuracy on the task}
\begin{figure}[h]
\centering
\includegraphics[height=120px]{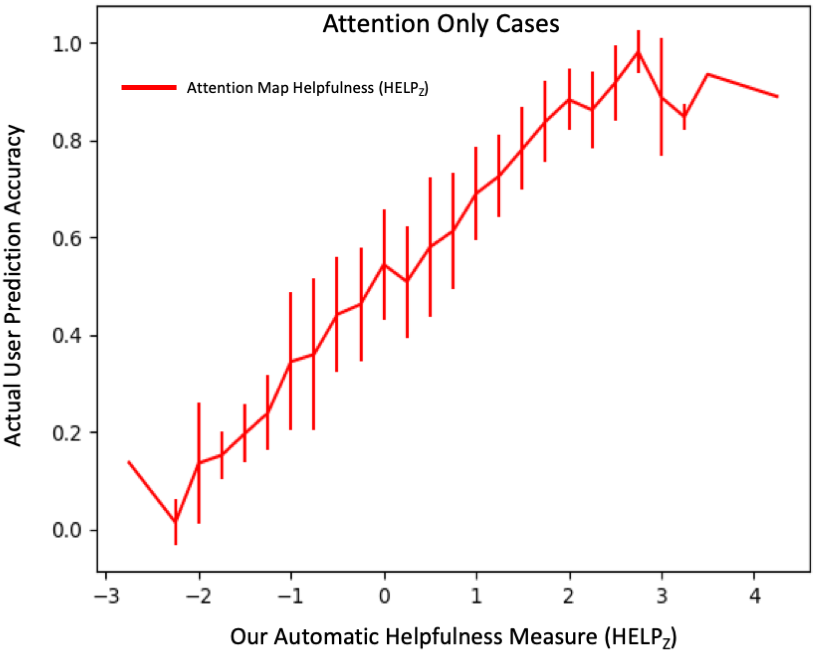}
\caption{Users' accuracy in predicting model correctness gets better on subsets that have higher helpfulness scores as evaluated by our automated metric.} 
\label{fig:metric_eval_attonly}
\end{figure}
As illustrated in Figure \ref{fig:metric_eval_attonly}, we note that there is a strong positive correlation between our $\helpsmall^A_Z$ score and actual users' accuracy in predicting model correctness. To compute the graph in Figure \ref{fig:metric_eval_attonly}, we sample subsets of our user study set of varying user accuracies and compute the helpfulness $\helpsmall_Z$ on those sets. We then plot the average users' accuracy in predicting model performance for these binned $\helpsmall_Z$ values.

\begin{figure*}[t]
\centering
\includegraphics[height=120px]{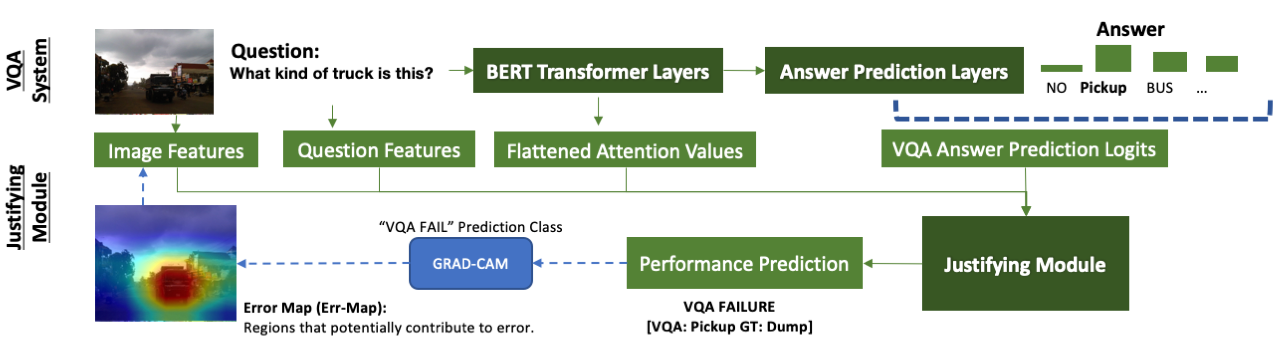}
\caption{We train a \inspector\ Module that learns to predict VQA System's performance. We compute the visual evidence for VQA Failure from the \inspector\ Module that highlights the error-contributing areas in the image for the given question (referred to as \errorcam). In the example shown, the question is ``What kind of truck is this?'', the \errorcam\ regions point to the truck, indicating that it might confuse the system. The VQA understandably answers incorrectly.} 
\label{fig:vqa_rm}
\end{figure*}

\xhdr{Helpfulness of Error Maps:} Using the same intuition as for attention maps, we can assume that users would also use a relevance-based criteria to judge error maps. However, since error maps point to error contributing areas, the helpfulness metric should be higher if their relevance is lower for cases where the VQA model produces a correct answers than it is for cases where the VQA model produces incorrect answers. 

To account for such an inverted behavior, we simply add a negative sign to indicate that a negative correlation is more helpful in the correlation based helpfulness:

\begin{align}
        \helpsmall^{ERR}_Z &= - ztest(\relsmall^{ERR}_c, \relsmall^{ERR}_w)
    \end{align}

With these definitions, a higher score in any $\helpsmall$ metric always means that the explanation is more helpful.

\xhdr{Helpfulness of Joint Error \& Attention Maps:}
Up to this point, our helpfulness scores rely on human attention maps being available
to determine the relevance of an explanation.
We also propose an automatic helpfulness metric that \textbf{does not require human annotated attention} when both attention and error maps are shown together, by computing the relevance (Spearman correlation) between the Error Map and Attention Map $\relsmall^{EA}$. 
As shown in Figure \ref{fig:atten_help_grid}c, we hypothesize that a joint error and attention map explanation is helpful if the relevance of the error map to the attention map is higher for cases where the VQA model produces wrong answers than for cases where the VQA model produces correct answers. Intuitively, this means that the explanation is helpful when the error-prone regions point to the attention regions for cases when the VQA produces an incorrect answer, and vice versa. Consequently, we can helpfulness ($\helpsmall^{EA}_Z$): 
\begin{align}
        \helpsmall^{EA}_Z &= - ztest(\relsmall^{EA}_c, \relsmall^{EA}_w)
\end{align}

\section{Explanation Generation}
\changed{In this section, we outline how we generate our error maps to improve helpfulness. We also outline a few ways we optimize our attention map visualization to maximize the proxy helpfulness characteristics.}

\subsection{Error Map Generation}
\changed{
As shown in the right side of Figure \ref{fig:vqa_rm}, we augment our VQA Model by a \inspector\ Module that learns to predict whether the VQA model will fail or not.}
 
 The Justifying Module is conditioned on the input image features $H_m$, word2vec \cite{mikolov2013distributed} question features~\footnote{word2vec~\cite{mikolov2013distributed} features are averaged across all question words}, \bertattn\ weights $W$, and the model's answer logits. We encode the image using a convolution layer and a fully connected layer into a 96-dim vector. The question features, \bertattn\ weights and answer logits are each encoded to 96 dimension hidden vectors using two fully connected layers (96 neurons each). We concatenate the four 96-dim vectors to form a latent vector $h_j$ that describes the context needed to predict the correctness.
 
We predict the \vqasystem's performance from the latent embedding $h_j$. We use a fully connected layer on $h_j$ followed by a sigmoid activation to output a binary probability for failure (output is 1 when VQA-System fails). Next, we use GradCAM \cite{selvaraju2017grad}, which uses gradients to highlight input regions that are important for a failure output. Specifically, we apply it to the logit of the failure predictor output to highlight regions in the image that contribute to the \inspector\ Module predicting \emph{Failure}. We use the gradient with respect to the image features $H_m$ to compute the Error Map.
Note that this error map geenration requires \textbf{no extra labels} other than the VQA dataset labels used to train the VQA model. 

\subsection{Finding Helpful Attention Maps}
\label{sec:bertexpl}
\changed{To produce a 2-D attention map from \bertattn{} layers, we previously described a hand-crafted way to generate attention maps in Section \ref{sec:vqanet}. Here, we outline a few ways to optimize the visualization by utilizing our \helpsmall metrics.}

\xhdr{Most Helpful \bertattn\ Attention (BestBERT) } 
We first extract heat maps from the attention weights $W$ for each layer and head. For layer $l$ and head $h$ the corresponding attention map over the $7 \times 7= 49$ image features is $A^{l,h} = \frac{1}{115} \sum_i W_{l, h, i, :49}$.
We compute the $\helpsmall^A_Z$ scores of attentions from each \bertattn\ layer and head, $A^{l,h}$ on a held out val split. 
We choose the $A^{l,h}$ with the highest $\helpsmall^A_Z$ (referred to as \textbf{BestBERT}). For completeness, we also generate attention by choosing the best head (and average the attention values over the layers for that head), referred to as \textbf{BestBERT-H}. Similarly, we also choose the best layer (and average attention over the heads), referred to as \textbf{BestBERT-L}.

\xhdr{Justified Attention Generation (\refattsmall)}
We also learn to predict human annotated attention~\cite{vqahat} from the Justifying Module's embedding, $h_j$, by tacking on another head to the module in addition to the failure prediction. This is similar to the work on generating justifying statements \cite{hendricks2016generating}, but done for attention maps. 
The embedding $h_j$ is fed through a fully connected layer with a 49 dimensional sigmoid output and then reshaped into a $7 \times 7$ attention map.
The Justifying Attention and Performance Prediction heads are trained simultaneously with a joint loss.
The first term of the loss minimizes the mean squared error between the predicted {\refattsmall}
 attention map and the corresponding human attention map~\cite{vqahat}.
The second term is binary cross entropy for failure prediction.

\section{Experimental Settings}
\changed{We first evaluate our new attention maps and error maps on our automated proxy helpfulness metrics.  
We then conduct user-studies as a gold standard evaluation to verify the findings from our automated metrics and show improved user performance when using our attention + error map explanations. 
Using this user study, we also test whether our automated helpfulness metrics for attention and error maps actually are indicative of human performance at predicting model correctness.
The test split of the Human Attention Dataset \cite{vqahat} is used for both automatic evaluation and user studies.
It consists of 3120 IQ pairs. For choosing hyper-parameters and checkpoints, we use a held-out val set of 1000 IQ pairs separate from the 3120 test set.}

\xhdr{Metrics Shown}
\changed{In the tables for automated metric-based evaluation, we show the following metrics as defined in Section \ref{sec:helpfulness}.}
\begin{compactenum}[--]
\item \textbf{Relevance \checkmark{} / Relevance $\times$}: The average relevance of the explanation mode for cases when the system is correct (Relevance \checkmark) or incorrect (Relevance $\times$). Hence, it is the average value of $\relsmall^A$ for attention maps, $\relsmall^{ERR}$ for error maps, and $\relsmall^{EA}$ for joint attention-error map explanations.
\item \textbf{\helpsmall$_{Z}$}: Helpfulness of the mode of explanation measured by the z-test approach as described in Section \ref{sec:helpfulness_approach}.  
\end{compactenum}

\xhdr{Baselines}
\changed{We compare our new attention maps to the current standard for generating them - \bertattn\ as described in Section \ref{sec:vqanet}.
To create a baseline error map based on \bertattn{}, we use the error map version of helpfulness, producing BertErr by choosing the attention $A^{l,h}$ that maximizes $\helpsmall^{ERR}_Z$ over all layers and heads.}

\xhdr{User Studies}
Users were divided into four groups - no explanations group (None), a group that saw BaselineBERT attention maps (described in Section \ref{sec:helpfulness}), a group that saw BestBERT Attention Maps + \errorcam\ Error Maps, and a group that saw \refattsmall\ + \errorcam. Workers in each group were kept separate and five independent workers predicted the performance on each IQ pair in each group to ensure the quality of our user studies. In total, we have 3300 ratings for the ``None'' group, 3000 for BaselineBERT, 2250 for BestBERT + \errorcam , and 3960 for \refattsmall\ + \errorcam.    

\begin{table}[t]
\centering
\caption{The \relevance\ and \usability\ of attention and error maps. Higher \usability\ is better. * indicates a statistically significant Z-test t-statistic score (Z) for $\helpsmall_Z$ at 99\% confidence level $p<0.01$.}
\vspace{1mm}
\label{table:attention_correlations}
\begin{tabular}{lcccc}
\toprule
\multirow{2}{*}{\textbf{\begin{tabular}[c]{@{}c@{}}Attention/Error \\ Maps\end{tabular}}} & \multicolumn{2}{c}{\textbf{Relevance}} & \textbf{$\helpsmall_Z$} $\uparrow$ \\
 & \checkmark & $\times$ & \\  \midrule
\footnotesize{a. }\textbf{BaselineBERT} & 0.32 & 0.31 & 0.4 \\ \midrule
\footnotesize{b. }\textbf{BestBERT-H} & 0.33 & 0.32 & 1.07 \\
\footnotesize{c. }\textbf{BestBERT-L} & 0.41 & 0.40 & 0.98 \\
\footnotesize{d. }\textbf{\refattsmall } & 0.57 & 0.54 & \textbf{2.56*} \\ 
\footnotesize{e. }\textbf{BestBERT} & 0.35 & 0.31 & \textbf{3.87*} \\ 
\midrule
\footnotesize{f. }\textbf{BERTErr} & 0.208 & 0.227 & 1.52 \\
\footnotesize{g. }\textbf{\errorcam } & 0.209 & 0.28 & \textbf{4.93*} \\
\bottomrule
\end{tabular}
\end{table}

\section{Results and Analysis}

We first establish the performance of the proposed explanations according to our automated metrics,
showing that our attention maps (e.g., \refattsmall, BestBERT) and error map (\errorcam) score higher for helpfulness than baselines. We then conduct user studies to show that explanations with higher helpfulness scores indeed improve users' ability to predict model performance. Below, we describe our key takeaways. 

\subsection{Automatic Evaluation of Explanations}

\xhdr{Our Error Map has high \usability\ }
Helpfulness scores for attention and error maps are listed in Table \ref{table:attention_correlations}. Our error map, \errorcam\ (row g), has the highest $\helpsmall_z$ score when compared to all attention maps (rows a to e) and baseline error map (row f). This is reflected in the difference in relevance for system correct vs wrong cases, which is the highest in \errorcam{}.

\xhdr{Our new attention maps are also more helpful than baselines, with BestBERT being the most helpful.}
Our relevance and helpfulness metrics for attention maps are shown in Table \ref{table:attention_correlations}. 
The proposed attention maps, \refattsmall\ (row d), BestBERT (row e), BestBERTHead (row b), and BestBERTLayer (row c) all show higher \usability{} compared to the baseline (row b). 
This is reflected in the relevance values, where a larger gap between correct and incorrect relevance leads to higher \usability{}. BestBERT has the highest difference between the correct and wrong cases' relevance, and the highest \usability{}.

\xhdr{Justifying Attention has high relevance, but relatively low helpfulness.}
BestBERT outperforms \refattsmall, which is supervisedly trained to be human like.
This shows that when we train attentions to be human-like, they seem more relevant (points to areas a human would point to for the question), but they may not be helpful for understanding model performance because they are more likely to be relevant even when the model is incorrect.

\xhdr{Error Maps used in tandem with attention maps improve helpfulness.}
In Table \ref{table:attentionerror_correlations},
we observe that the $\helpsmall^{EA}$ scores of attention maps with our ErrMap are higher the attention map's $\helpsmall^A$ scores. This is especially true for high relevance, yet low helpfulness attention maps (column $\helpsmall^A$ vs $\helpsmall^{EA}$, Table \ref{table:attentionerror_correlations}). 
For example, BaselineBERT and \refattsmall\, which have high relevance even for wrong answers are not very helpful (row a, e, same as Table \ref{table:attention_correlations} numbers). When combined with \errorcam\, the joint helpfulness increases significantly. 
This is because our ErrMap points to relevant areas when the model errs, and hence, the alignment with an always-relevant attention map can easily indicate failure. 

\begin{table}[t]
\centering
\caption{The \relevance\ and \usability\ of explanations when various attentions maps are used in conjunction with \errorcam. Higher \usability\ is better. * indicates a statistically significant Z-test t-statistic score at 99\% confidence level $p<0.01$}
\vspace{1mm}
\label{table:attentionerror_correlations}
\begin{tabular}{lcc} \toprule
\multicolumn{1}{c}{\multirow{2}{*}{\textbf{\begin{tabular}[c]{@{}c@{}}Attn \\ Choice\end{tabular}}}} & \multicolumn{2}{c}{\textbf{Helpfulness}} \\
\multicolumn{1}{c}{} & \textbf{$\helpsmall^A$} & \textbf{$\helpsmall^{EA}$} \\ \midrule
a. \textbf{BaselineBERT} & 0.4 & \textbf{5.94} \\
b. \textbf{BestBERT-H} & 1.07 & \textbf{4.59} \\
c. \textbf{BestBERT-L} & 0.98 & \textbf{7.15} \\
d. \textbf{BestBERT} & 2.56 & \textbf{4.99} \\
e. \textbf{J-Att} & 3.87 & \textbf{7.71} \\ \bottomrule
\end{tabular}
\end{table}

\subsection{User-based Evaluation}

\begin{figure}[t]
\centering
\includegraphics[width=0.85\textwidth]{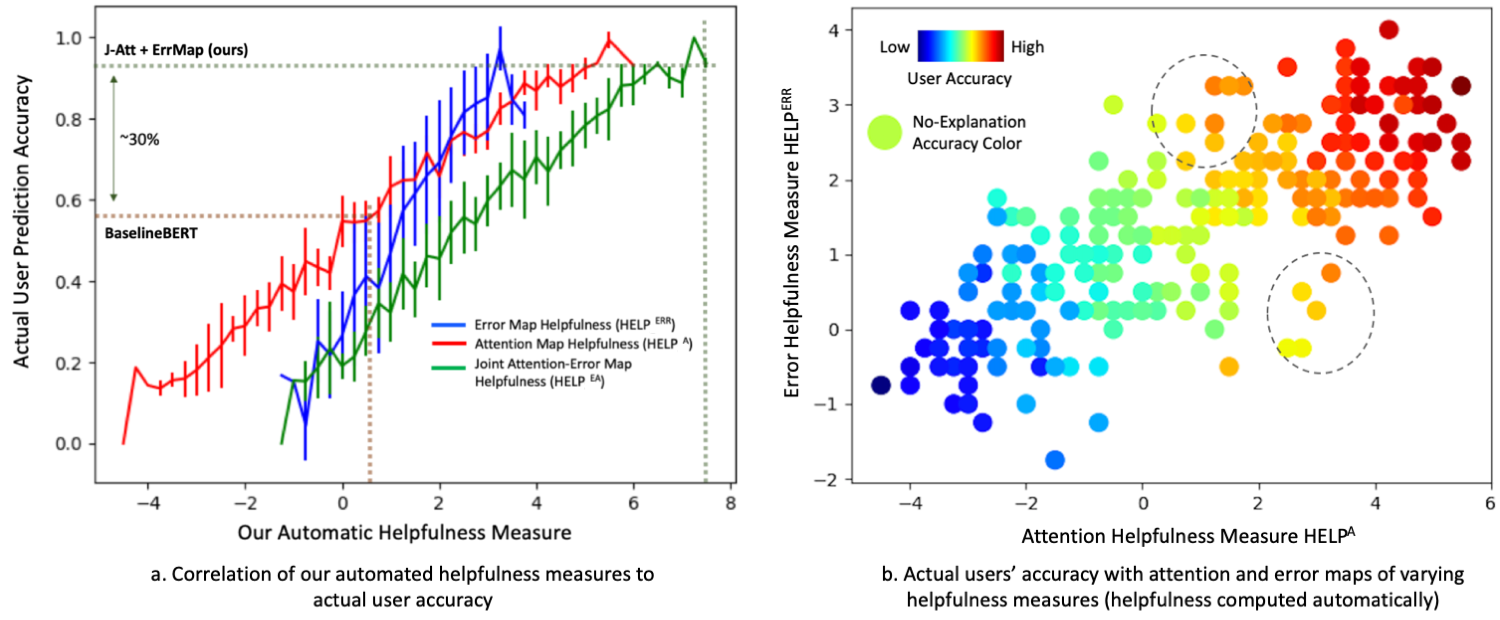}
\caption{\textbf{a.} Our automatic metrics for evaluating the \usability\ of attention and error maps correlates to and can reflect users' ability to predict model performance.
\textbf{b.} The distribution of user prediction accuracy across varying $\helpsmall^A_Z$ of attention maps and $\helpsmall^{ERR}_Z$ of error maps. Having either helpful Attention or Error maps can improve user accuracy as highlighted by the off diagonal dashed circles.} 
\label{fig:metric_eval}
\end{figure}

We first evaluate our proxy helpfulness metrics for error maps and joint error+attention maps and check whether they correlate to actual human accuracy on the task. Similar to how we tested whether our attention helpfulness metric in Figure \ref{fig:metric_eval_attonly}, we plot the average user accuracy for sets of IQ pairs with varying $\helpsmall^{ERR}_Z$ and $\helpsmall^{EA}_Z$ values in Fig.~\ref{fig:metric_eval} . 

\xhdr{Our automated \usability\ metrics for error maps and joint error+attention maps correlate with users' accuracy to predict model performance}
As visualized in Figure \ref{fig:metric_eval}a, we see a strong positive correlation (Pearson Correlation $>0.9$) with $\helpsmall_Z$ scores for attention and error maps to the user prediction accuracy when using those explanations. 
The red line shows the attention only $\helpsmall^A_Z$, the blue line shows the $\helpsmall^{ERR}_Z$, and the green line shows the $\helpsmall^{EA}_Z$. 

\xhdr{Our explanations improve user prediction accuracy}
Table \ref{table:predtask_acc} shows the users' overall prediction accuracy (AVG), accuracy for cases with the VQA system producing correct answers (\checkmark), and accuracy for cases with the VQA system producing wrong answers ($\times$). 
We observe an increase in overall accuracy using \errorcam\ along with BestBERT (row c) or \refattsmall\ (row d) explanations than when using no explanations (row a) or when using Baseline BERT explanations (row b). Note that the \helpsmall\ score on the set used in the user study is lower than that on overall val data. 
Hence, we also mark the user accuracy corresponding to the set that has the helpfulness scores closest to our overall test data using dotted lines for baselineBERT (attention-only) and our best performing mode, \refattsmall\ + \errorcam\ (joint attention-error). We note a $30\%$ improvement in user accuracy (Fig.~\ref{fig:metric_eval}a). 

\xhdr{Our explanations help especially in predicting failure}
We note an increase of $11\%$ (Table \ref{table:predtask_acc} comparing rows a and c in column $\times$) in user accuracy to predict failure of the VQA system.


\begin{table}[]
\centering
\caption{User accuracy on our user-study set (randomly sampled subset of val data) for predicting \vqasystem\ performance. Subset \helpsmall\ indicates the \helpsmall\ score on the user study set used. }
\label{table:predtask_acc}
\vspace{1mm}
\begin{tabular}{lccll} \toprule
\multirow{2}{*}{\textbf{Expl. Mode}} & \multirow{2}{*}{\textbf{\begin{tabular}[c]{@{}c@{}}Subset\\ \helpsmall\end{tabular}}} & \multicolumn{3}{c}{\textbf{User Pred Acc}} \\ \cline{3-5}
 &  & \textbf{\small{AVG}} & \multicolumn{1}{c}{\textbf{\checkmark}} & \multicolumn{1}{c}{\textbf{$\times$}} \\ \midrule
\footnotesize{a. }\textbf{None} & -- & 57.18 & 64.02 & 49.31 \\
\footnotesize{b. }\textbf{BaselineBERT} & 0.57 & 56.87 & 63.16 & 49.56 \\
\footnotesize{c. }\textbf{BestBERT+ErrMap} & 2.54 & 60.00 & 59.53 & \textbf{60.42} \\
\footnotesize{d. }\textbf{J-Att + ErrMap} & \textbf{4.37} & \textbf{60.13} & \textbf{65.21} & 55.06 \\ \bottomrule
\end{tabular}
\end{table}

\xhdr{Gameability of the automated metric}
We discuss a few contrived explanations that may not reflect the model's logic for processing the QI pair, but can score high on our metric. 
\begin{compactenum}[--]
\item We generate a centered gaussian attention for all images. We note that this has reasonably high helpfulness ($\helpsmall^A_Z = 2.35$). However, this is indeed a helpful explanation about VQA behavior since a high $\helpsmall^A_Z$ indicates that VQA models tend to answer correctly when the required objects are in the centered attention region.  
\item As a strong upperbound, we generate centered attentions when our \inspector\ Module predicts the system will be correct and randomized attention otherwise. Since the \inspector\ module is reasonably accurate at predicting VQA performance ($70\%$ accurate), attention is relevant for correct answers, and random (low relevance) for incorrect answers. Hence, this attention  has high helpfulness ($\helpsmall^A_Z=14.51$) since it gives a clear visual indicator of VQA correctness (i.e., the ground truth for the user prediction task).
\end{compactenum}
Note that these explanations are gamed because they are not faithful to the internal workings of the model. However, the gamed explanations can indeed provide some helpful insights to users by revealing certain biases.



\xhdr{Having either helpful Attention or Error Maps is beneficial for predicting VQA performance}
In Figure \ref{fig:metric_eval}b, we plot the user study results with respect to the $\helpsmall^A_Z$ of attention maps along the x-axis and the $\helpsmall^{ERR}_Z$ of error maps along the y-axis. The color of the dots indicate the user prediction accuracy (blue is lower and red is higher). We also show the color corresponding to the accuracy for users seeing no explanations for comparison. While it is expected that the user accuracy is higher when both attention and error maps are more helpful, we note that even having either one as helpful can have reasonably high user accuracy as highlighted by the off-diagonal circles.

\section{Related Works}
Historically, explainable AI (XAI) spanned from medical systems \cite{shortliffe1984model} to using AI for educational purposes \cite{lane2005explainable,van2004explainable}. There has been recent thrust in developing XAI systems \cite{gunning2019darpa,adadi2018peeking} for deep networks and a major goal has been to improve users' mental model of systems for better human-AI collaboration \cite{chandrasekaran2017takes,chandrasekaran2018explanations, hendricks2016generating, ray2019can, alipour2020study} or discover biases in models \cite{manjunatha2019explicit}. However, most of these works show no significant improvement of users' accuracy to predict performance with attention maps or show somewhat deleterious effects \cite{poursabzi2018manipulating}. In this paper, we argue that to design helpful explanations, we need to have a modeling of users' mental model of AI (perception of AI output). We use that to develop helpful attention and error maps to improve users' ability to predict the model performance. 

There have also been works on modeling AI to understand a human's mental model - their actions, motivations or mental states \cite{vondrick2016anticipating, vondrick2016predicting, el2005real}, human interactions \cite{oliver2000bayesian}, or anticipating their beliefs \cite{eysenbach2017mistaken}. Systems have also been trained to learn how to behave in society \cite{chuang2018learning} for better human-AI collaboration \cite{puig2020watch}. While they mostly focus on how to model human behavior or interactions, we focus on modeling how humans interpret system outputs to aid their mental model of the AI system.  

We use Visual Question Answering (VQA) \cite{antol2015vqa} as our choice of AI system. Most effective approaches use attention to weigh image and text features \cite{pythia18arxiv,lu2016hierarchical,teney2017tips,xu2016ask,kazemi2017show}. Recently, transformer-based \cite{vaswani2017attention} attention is used \cite{lu2019vilbert, bertsanvqa, li2019visualbert, alipour2020impact}.
However, most of these works do not specify how to effectively visualize the high dimensional transformer attentions. We utilize our automated helpfulness metrics to compute effective ways of displaying attention. There are also works on improving saliency-based explanations by making them more faithful \cite{petsiuk2018rise, chang2018explaining, bargal2018guided}, using model introspection \cite{chang2018explaining}, grounding to evidence \cite{hendricks2018grounding}, or making them human-like \cite{qiao2017exploring}. However, we focus on how to design explanations to not just be relevant, but also be helpful to human users.

To improve helpfulness, we also introduce a error maps for pointing to error-prone regions for VQA. Predicting failures have also been proposed earlier on other domains for better human understanding \cite{zhang2014predicting, bansal2014towards, mcgrath2020does} and to improve models \cite{patro2019u}. This supports our motivation for introducing error maps in tandem with attention explanations to improve users' ability to predict model performance for VQA.

\section{Conclusions}
In this paper, we proposed Error Maps in VQA as a novel explanation method that improve helpfulness over just showing attention maps. We also revealed characteristics of heatmap-style explanations that make them helpful to users and designed an automated proxy helpfulness metric based on it.  
We anticipate that our automated metrics can aid rapid development of heatmap explanations without having to conduct repeated user-studies, and that our proposed Error Map explanations can be applied to other AI domains to improve users' mental model of those systems for better human-AI collaboration.

\section*{Acknowledgements}
 We would like to thank Anirban Roy, Briland Hitaj and Karan Sikka for their comments and feedback. This research was developed with funding from DARPA under the \reva{Explainable AI (XAI)} program. The views, opinions and/or findings expressed are those of the authors’ and should not be interpreted as representing the official views or policies of the Department of Defense or the U.S. Government. \footnote{Authors have no conflict of interest}

\bibliography{wileyNJD-AMA}%




\end{document}